\documentclass[11pt,a4paper]{article}

\usepackage{microtype}

\usepackage[hyperref]{emnlp2018}
\usepackage{times}
\usepackage{latexsym}
\usepackage{multirow}
\usepackage{todonotes}
\usepackage{url}
\usepackage[utf8]{inputenc}

\usepackage{wrapfig}
\usepackage{lineno}
\usepackage[normalem]{ulem}
\usepackage{natbib}
\usepackage{graphicx}
\usepackage{booktabs}

\usepackage{tikz}
\usepackage{bm}
\usepackage{subfigure} 
\usepackage{xcolor,colortbl}
\usepackage{algorithmicx}

\usepackage[normalem]{ulem}

\usetikzlibrary{chains, positioning}

\definecolor{Gray}{gray}{0.85}
\definecolor{yellow}{rgb}{1,1,0.5}

\modulolinenumbers[5]

\aclfinalcopy 

\title{UAlacant machine translation quality estimation at WMT 2018: a simple approach using phrase tables and feed-forward neural networks}

\author{Miquel Esplà-Gomis \,\,\,\, Felipe Sánchez-Martínez \,\,\,\, Mikel L.\ Forcada\\
  Departament de Llenguatges i Sistemes Informàtics \\
  Universitat d'Alacant,  E-03071 Alacant, Spain\\
  \texttt{\{mespla,fsanchez,mlf\}@dlsi.ua.es} }

\date{}

\begin{document}
\maketitle
\begin{abstract}
 We describe the Universitat d'Alacant submissions to the word- and sentence-level machine translation (MT) quality estimation (QE) shared task at WMT 2018. Our approach to word-level MT QE builds on previous work to mark the words in the machine-translated sentence as \textit{OK} or \textit{BAD}, and is extended to determine if a word or sequence of words need to be inserted in the gap after each word. Our sentence-level submission simply uses the edit operations predicted by the word-level approach to approximate TER. The method presented ranked first in the sub-task of identifying insertions in gaps for three out of the six datasets, and second in the rest of them.
\end{abstract}

\section{Introduction} \label{se:introduction}

This paper describes the Universitat d'Alacant submissions to the word- and sentence-level machine translation (MT) quality estimation (QE) shared task at WMT 2018~\citep{WMT2018}. Our approach is an extension of a previous approach \citep{espla-gomis15wmt,espla-gomis15eamt,espla2016ualacant} in which we simply marked the words \(t_j\) of a machine-translated segment \(T\) as \textit{OK} (no changes are needed) or as \textit{BAD} (needing editing). Now we also mark the gaps \(\gamma_j\) after each word \(t_j\) as \textit{OK} (no insertions are needed) or as \textit{BAD} (needing the insertion of one or more words). In addition, we use the edit operations predicted at the word level to estimate quality at the sentence level.

The paper is organized as follows: section~\ref{se:prevwork} briefly
reviews previous work on word-level MT QE; section~\ref{se:method} describes the method used to label words and gaps, paying special attention to the features extracted (sections~\ref{ss:featwords} and \ref{ss:featpositions}) and the neural network (NN) architecture and
its training (section~\ref{ss:architecture}); section~\ref{ss:experiments} describes the datasets used; section~\ref{se:results} shows the main results; and, finally, section~\ref{se:conclusion} closes the paper with concluding remarks.

\section{Related work} \label{se:prevwork}

Pioneering work on word-level MT QE dealt with predictive/interactive MT \citep{gandrabur03,blatz04,ueffing05,ueffing07}, often under the name of \emph{confidence estimation}. Estimations relied on the internals of the actual MT system ---for instance, studying the $n$-best translations \citep{ueffing07}--- or  used external sources of bilingual information; for instance, both \citet{blatz04} and \citet{ueffing05} used probabilistic dictionaries; in the case of \citet{blatz04}, as one of many features in a binary classifier for each word.

The last decade has witnessed an explosion of work in word-level MT QE, with most of the recent advances made by participants in the shared tasks on MT QE at the different editions of the Conference on Statistical Machine Translation (WMT). Therefore, we briefly review those papers related to our approach: those using an external bilingual source such as an MT system and those using NN.

As regards work using \emph{external bilingual resources}, we can highlight four groups of contributions:
\begin{itemize} \setlength{\itemsep}{-2pt}
\item To estimate the sentence-level quality of MT output for a source
  segment \(S\),
  \citet{bicici13} chooses sentence pairs from a parallel corpus 
  which are close to \(S\), and builds an SMT system whose internals when
  translating \(S\) are examined to extract features.
\item MULTILIZER, one of the participants in the sentence-level
  MT QE task at WMT~2014~\citep{WMT14} uses other MT
  systems to translate \(S\) into the target language (TL) and \(T\)
  into the source language (SL). The results are compared to the original
  SL and TL segments to obtain indicators of quality.
\item \citet{blain2017bilexical} use \emph{bilexical embeddings}
  (obtained from SL and TL word embeddings and word-aligned parallel
  corpora) to model the strength of the relationship between SL and TL
  words, in order to estimate sentence-level and word-level MT quality.
\item Finally,
  \citet{espla-gomis15wmt,espla-gomis15eamt}, and \citet{espla2016ualacant}
  perform word-level MT QE by using other MT systems to translate sub-segments
  of \(S\) and \(T\)
  and extracting features describing the way in which these translated
  sub-segments match sub-segments of \(T\).
  This is the work most related to the one presented in this paper.
\end{itemize}
Only the last two groups of work actually tackle the problem of \emph{word-level} MT QE, and none of them are able to identify the gaps where insertions are needed.

As regards the use of \emph{neural networks} (NN) in MT QE, we can highlight a few contributions:
\begin{itemize}\setlength{\itemsep}{-2pt}
\item \citet{kreutzer2015quality} use a deep feed-forward NN to
  process the concatenated vector embeddings of neighbouring TL words
  and (word-aligned) SL words into feature vectors ---extended with
  the baseline features provided by WMT15~\citep{WMT15} organizers--- to perform word-level MT QE.
\item \citet{martins16} achieved the best results in the word-level MT QE
  shared task at WMT 2016~\citep{WMT16} by combining a feed-forward NN
  with two recurrent NNs whose predictions were fed into a linear
  sequential model together with the baseline features provided by the
  organizers of the task. An extension \citep{martins2017pushing} uses the
  output of an automatic post-editing tool, with a clear improvement
  in performance.
\item \citet{kim2017predictor,kim2017predictor2} obtained in WMT
  2017~\citep{WMT17} results which were better or comparable to those by
  \citet{martins2017pushing}, using a three-level stacked architecture
  trained in a multi-task fashion, combining a neural word prediction
  model trained on large-scale parallel corpora, and word- and sentence-level MT QE models.
\end{itemize}
Our approach uses a much simpler architecture than the last two
approaches, containing no recurrent NNs, but just 
feed-forward NNs applied to a fixed-length context window
around the word or gap about which a decision is being
made (similarly to a convolutional approach). This makes our approach
easier to train and parallelize.

\section{Method} \label{se:method}

The approach presented here builds on previous work by the same
authors \citep{espla-gomis15wmt,espla-gomis15eamt,espla2016ualacant} in which insertion
positions were not yet predicted and a slightly different feature set was
used. As in the original papers, here we use black-box bilingual
resources from the Internet. In particular, we use, for each language pair,
the statistical MT phrase
tables available at OPUS\footnote{\protect\url{http://opus.nlpl.eu/}}
to spot sub-segment correspondences between the SL segment \(S\)
and its machine translation \(T\)
into the TL (see section~\ref{ss:OPUS} for details). This is done by
dividing both \(S\)
and \(T\)
into all possible (overlapping) sub-segments, or $n$-grams, up to a
certain maximum length.\footnote{For our submission, we used $L=5$.} These sub-segments are then translated into
the TL and the SL, respectively, by means of the phrase tables
mentioned (lowercasing of sub-segments before and after translation is
used to increase the chance of a match). These sub-segment
correspondences are then used to extract several sets of
features that are fed to a feed-forward NN in order to
label the words and the gaps between words as \textit{OK} or as \textit{BAD}. One of the main advantages of this approach,
when compared to the other approaches described below, is that it uses
simple string-level bilingual information extracted from a publicly
available source to build features that allow us to easily estimate
quality  for the words and inter-word gaps in \(T\).

\subsection{Features for word deletions} \label{ss:featwords}
We define three sets of features to detect the words to be deleted: one taking advantage of the sub-segments \(\tau\) that appear in \(T\), \(\mathrm{Keep}_n(\cdot)\); another one that uses the translation frequency with which a sub-segment \(\sigma\) in \(S\) is translated as the sub-segment \(\tau\) in \(T\), \(\mathrm{Freq}_n^\mathrm{keep}(\cdot)\); and a third one that uses the alignment information between \(T\) and \(\tau\) and which does not require \(\tau\) to appear as a contiguous sub-segment in \(T\), \(\mathrm{Align}_n^\mathrm{keep}(\cdot)\).

\paragraph{Features for word deletions based on sub-segment pair occurrences (\(\mathrm{Keep}\))}\label{subsub:keep}


Given a set of sub-segment pairs \(M = \{(\sigma,\tau)\}\) coming from the union of several phrase tables,
the first set of features, \(\mathrm{Keep}_n(\cdot)\), is obtained by computing the amount of sub-segment translations \((\sigma,\tau) \in M\) with \(|\tau|=n\) that confirm that word \(t_j\) in \(T\) should be kept in the translation of \(S\). A sub-segment translation \((\sigma,\tau)\) confirms \(t_j\) if \(\sigma\) is a sub-segment of \(S\), and \(\tau\) is an \(n\)-word sub-segment of \(T\) that covers position \(j\). This set of features is defined as follows:
\begin{displaymath}
\begin{array}{r} \mathrm{Keep}_n(j,S,T,M) = \\ = \displaystyle{\frac{|\{\tau: (\sigma,\tau) \in \mathrm{conf}_n^\mathrm{keep}(j,S,T,M)\}|}{|\{\tau: \tau \in \mathrm{seg}_{n}(T) \mbox{\,} \land \mbox{\,} j \in \mathrm{span}(\tau,T)\}|}}\end{array}
\end{displaymath}
where \(\mathrm{seg}_n(X)\) represents the set of all possible \(n\)-word sub-segments of segment \(X\), and function \(\mathrm{span}(\tau,T)\) returns the set of word positions spanned by the sub-segment \(\tau\) in the segment \(T\); if \(\tau\) is found more than once in \(T\), it returns all the possible positions spanned. Function \(\mathrm{conf}_n^\mathrm{keep}(j, S, T, M)\) returns the collection of sub-segment pairs \((\sigma,\tau)\) that confirm a given word \(t_j\), and is defined as:
\begin{displaymath}
  \nonumber \begin{array}{r}\mathrm{conf}_n^\mathrm{keep}(j,S,T,M) = \\ = 
  \{(\sigma,\tau)\in \mathrm{match}_n(M,S,T) : \mbox{\,} j \in \mathrm{span}(\tau,T)\} \end{array}
\end{displaymath}
where 
\(\mathrm{match}_n(M,S,T))\)
is the set of phrase pairs \((\sigma,\tau) \in M\)  such that 
\(\sigma\in\mathrm{seg}_*(S)\) and \(\tau\in \mathrm{seg}_n(T)\), 
and where \(\mathrm{seg}_*(S)=\cup_{n=1}^{\infty} \mathrm{seg}_n(S)\).\footnote{\citet{espla-gomis15wmt} conclude that constraining only the length of \(\tau\) leads to better results than constraining both \(\sigma\) and \(\tau\).}

\paragraph{Features for word deletions based on sub-segment pair occurrences using translation frequency (\(\mathrm{Freq}_n^\mathrm{keep}\))}\label{subsub:keepfreq}

The second set of features uses the probabilities of subsegment pairs. 
To obtain these probabilities from a set of phrase tables, we first use the count of joint occurrences of \((\sigma,\tau)\) provided in each phrase table.
Then, when looking up a SL sub-segment $\sigma$, the probability $p(\tau|\sigma)$ is computed across all phrase tables from the accumulated counts. 
Finally, we define \(\mathrm{Freq}_n^\mathrm{keep}(\cdot)\) as:
\begin{displaymath}
\begin{array}{r} \mathrm{Freq}_n^\mathrm{keep}(j,S,T,M) = \\ =  \displaystyle{\sum_{(\sigma,\tau) \in \mathrm{conf}_n^\mathrm{keep}(j,S,T,M)}
p(\tau|\sigma). }\end{array}
\end{displaymath}


\paragraph{Features for word deletions based on word alignments of partial matches (\(\mathrm{Align}_n^\mathrm{keep}\))} \label{subsub:keepalign}

The third set of  features takes advantage of partial matches, that is, of sub-segment pairs \((\sigma, \tau)\) in which \(\tau\) does not appear as such in \(T\). 
This set of features is defined as:
\begin{equation}
\label{eq:aligkeep}
\begin{array}{r} \mathrm{Align}_n^\mathrm{keep}(j,S,T,M,e) = \\ = 
\displaystyle{\sum_{\tau \in \mathrm{segs\_edop}_n(j,S,T,M,e)} \frac{|\mathrm{LCS}(\tau, T)|}{|\tau|}}\end{array}
\end{equation}
where $\mathrm{LCS}(X,Y)$ returns the word-based longest common sub-sequence between segments $X$ and $Y$, and \(\mathrm{segs\_edop}_n(j,S,T,M,e)\) returns the set of sub-segments \(\tau\) of length \(n\) from \(M\) that are a translation of a sub-segment \(\sigma\) from \(S\) and in which, after computing the LCS with \(T\), the \(j\)-th word \(t_j\) is assigned the edit operation \(e\):\footnote{Note that the sequence of edit operations needed to transform $X$ in $Y$ is a by-product of computing $\mathrm{LCS}(X,Y)$; these operations are \texttt{insert}, \texttt{delete} or \texttt{match} (when the corresponding word does not need to be edited).}
\begin{equation}
\label{eq:segsedop}
\begin{array}{rcl} 
\mathrm{segs\_edop}_n(j,S,T,M,e) = \\ =  \{(\tau: (\sigma,\tau) \in M \land\; \sigma \in \mathrm{seg}_*(S) \\\land\; |\tau|=n \land\;  \mathrm{editop}(t_j, T, \tau) = e \} \end{array}
\end{equation}
where \(\mathrm{editop}(t_j, T, \tau)\) returns the edit operation assigned to \(t_j\) and \(e\) is either \(\mathtt{delete}\) or  \(\mathtt{match}\). If \(e=\mathtt{match}\) the resulting set of features provides evidence in favour of keeping the word \(t_j\) unedited, whereas when \(e=\mathtt{delete}\) it provides evidence in favour of removing it.
Note that features \(\mathrm{Align}_n^\mathrm{keep}(\cdot)\) are the only ones to provide explicit evidence that a word should be deleted. 

The three  sets of features described so far, \(\mathrm{Keep}_n(\cdot)\),
\(\mathrm{Freq}_n^\mathrm{keep}(\cdot)\), and \(\mathrm{Align}_n^\mathrm{keep}(\cdot)\), are computed for \(t_j\) for all the values of sub-segment length \(n \in [1,L]\). Features \(\mathrm{Keep}_n(\cdot)\) and \(\mathrm{Freq}_n^\mathrm{keep}(\cdot)\) are computed by querying the collection of sub-segment pairs \(M\) in both directions (SL--TL and TL--SL). 
Computing \(\mathrm{Align}_n^\mathrm{keep}(\cdot)\) only queries $M$ in one direction (SL--TL) but  is computed twice: once for the edit operation \texttt{match}, and once for the edit operation \texttt{delete}. 

\subsection{Features for insertion positions} \label{ss:featpositions}
In this section, we describe three sets of features ---based on those described in section~\ref{ss:featwords} for word deletions--- designed to detect insertion positions. The main difference between them is that the former apply to words, while the latter apply to gaps; we will call \(\gamma_j\) the gap after word \(t_j\).\footnote{Note that the index of the first word in \(T\) is 1, and gap $\gamma_0$ corresponds to the space before the first word in \(T\).}

\paragraph{Features for insertion positions based on sub-segment pair occurrences (\(\mathrm{NoInsert}\))}\label{subsub:noinsert}
The first set of features, \(\mathrm{NoInsert}_n(\cdot)\), based on the \(\mathrm{Keep}_n(\cdot)\) features for word deletions, is defined as follows:
\begin{displaymath}
\begin{array}{r} \mathrm{NoInsert}_n(j,S,T,M) = \\ \displaystyle{\frac{|\{\tau: (\sigma,\tau) \in \mathrm{conf}_n^{\mathrm{noins}}(j,S,T,M)\}|}{|\{\tau: \tau \in \mathrm{seg}_{n}(T) \mbox{\,} \land \mbox{\,} [j,j+1] \subseteq  \mathrm{span}(\tau,T)\}|}}\end{array}
\end{displaymath}
where function \(\mathrm{conf}_n^{\mathrm{noins}}(j, S, T, M)\) returns the collection of sub-segment pairs \((\sigma,\tau)\) covering a given gap \(\gamma_j\), and is defined as:
\begin{displaymath}
  \nonumber \begin{array}{r}\mathrm{conf}_n^\mathrm{noins}(j,S,T,M) = \\
  \{(\sigma,\tau)\in \mathrm{match}_n(M,S,T): \\{} [j,j+1] \subseteq  \mathrm{span}(\tau,T)\} \end{array}
\end{displaymath}
\(\mathrm{NoInsert}_n(\cdot)\) accounts for the number of times that the translation of sub-segment \(\sigma\) from \(S\) makes it possible to obtain a sub-segment \(\tau\) that covers the gap \(\gamma_j\), that is,  a \(\tau\) that covers both \(t_j\) and \(t_{j+1}\). If a word is missing in gap \(\gamma_j\), one would expect to find fewer sub-segments \(\tau\) that cover this gap, therefore obtaining low values for \(\mathrm{NoInsert}_n(\cdot)\), while if there are no words missing in $\gamma_j$, one would expect more sub-segments \(\tau\) to cover the gap, therefore obtaining values of \(\mathrm{NoInsert}_n(\cdot)\) closer to $1$. In order to be able to identify insertion positions before the first word or after the last word, we use imaginary sentence boundary words \(t_0\) and \(t_{|T|+1}\), which can also be matched,\footnote{These boundary words are annotated in \(M\) when this resource is built.} thus allowing us to obtain evidence for gaps \(\gamma_0\) and \(\gamma_{|T|}\).




\paragraph{Features for insertion positions based on sub-segment pair occurrences using translation frequency (\(\mathrm{Freq}_n^\mathrm{noins}\))}\label{subsub:noinsertfreq}
Analogously to \(\mathrm{Freq}_n^\mathrm{keep}(\cdot)\) above, we define the feature set \(\mathrm{Freq}^{\mathrm{noins}}_n(\cdot)\), now for gaps:
\begin{displaymath}
\begin{array}{r} \mathrm{Freq}^{\mathrm{noins}}_n(j,S,T,M) = \\[3mm] =  \displaystyle{\sum_{(\sigma,\tau) \in \mathrm{conf}_n^\mathrm{noins}(j,S,T,M)} p(\tau|\sigma)}\end{array}
\end{displaymath}



\paragraph{Features for insertion positions based on word alignments of partial matches (\(\mathrm{Align}_n^\mathrm{noins}\))} \label{subsub:noinsertalign}
Finally, the set of features \(\mathrm{Align}_n^\mathrm{keep}(\cdot)\) for word deletions can be easily repurposed to detect the need for insertions by setting the edit operation \(e\) in eq.~(\ref{eq:aligkeep}) to \(\mathtt{match}\) and \(\mathtt{insert}\) and redefining eq.~(\ref{eq:segsedop}) as
\begin{displaymath}
\begin{array}{r} 
\mathrm{segs\_edop}_n(j,S,T,M,e) =  \{\tau: (\sigma,\tau) \in M \\ \land\; \sigma\in\mathrm{seg}_*(S) \land\; |\tau|=n \\ \land \; \mathrm{editop}(t_j, \tau, T) = e \} \end{array}
\end{displaymath}
where the LCS is computed between \(\tau\) and \(T\), rather than the other way round.\footnote{It is worth noting that $\mathrm{LCS}(X,Y)=\mathrm{LCS}(Y,X)$, but the sequences of edit operations obtained as a by-product are different in each case.}  We shall refer to this last set of features for insertion positions as \(\mathrm{Align}_n^\mathrm{noins}(\cdot)\).






The sets of features for insertion positions, \(\mathrm{NoInsert}_n(\cdot)\),
\(\mathrm{Freq}_n^\mathrm{noins}(\cdot)\) and \(\mathrm{Align}_n^\mathrm{noins}(\cdot)\), are computed for gap \(\gamma_j\) for all the values of sub-segment length \(n \in [2,L]\). As in the case of the feature sets employed to detect deletions, the first two sets are computed by querying the set of subsegment pairs $M$ via the SL or via the TL, while the latter can only be computed by querying $M$ via the SL for the edit operations \texttt{insert} and \texttt{match}. 

\subsection{Neural network architecture and training} \label{ss:architecture}

We use a two-hidden-layer feed-forward NN to jointly predict the labels (\textit{OK} or \textit{BAD}) for word $t_j$ and gap $\gamma_i$, using features computed at word positions $t_{i-C}$, $t_{i-C+1},\ldots,$ $t_{i-1},$ $t_i$, $t_{i+1},\ldots,$ $t_{i+C-1},$ $t_{i+C}$ and at gaps $\gamma_{i-C},$ $\gamma_{i-C+1},\ldots,$ $\gamma_{i-1},$ $\gamma_i,$ $\gamma_{i+1},\ldots,$ $\gamma_{i+C-1},$ $\gamma_{i+C}$, where $C$ represents the amount of left and right context around the word and gap being predicted.

The NN architecture has a modular first layer with ReLU activation functions, in which the feature vectors for each word and gap, with $F$ and $G$ features respectively, are encoded into intermediate vector representations (``embeddings'') of the same size; word features are augmented with the baseline features provided by the organizers. The weights for this first layer are the same for all words and for all gaps (parameters are tied). A second layer of ReLU units combines these representations into a single representation of the same length $(2C+1)(F+G)$. Finally, two sigmoid neurons in the output indicate, respectively, if word $t_i$ has to be tagged as \textit{BAD}, or if  gap $\gamma_i$ should be labelled as \textit{BAD}. Preliminary experiments confirmed that predicting word and gap labels with the same NN led to better results than using two independent NNs.

The output of each of the sigmoid output units is additionally independently thresholded~\citep{Lipton2014} using a line search to  establish thresholds that optimize the product of the $F_1$ score for \textit{OK} and \textit{BAD} categories on the development sets. This is done since the product of the $F_1$ scores is the main metric of comparison of the shared task, but it cannot be directly used as the objective function of the training as it is not differentiable.

Training was carried out using the Adam stochastic gradient descent algorithm to optimize cross-entropy. A dropout regularization of 20\% was applied on each hidden layer. Training was stopped when results on the development set did not improve for 10 epochs. In addition, each network was trained 10 times with different uniform initializations~\citep{He:2015:DDR:2919332.2919814}, choosing the parameter set performing best on the development set.

Preliminary experiments have led us to choose a value $C=3$ for the number of words and gaps both to the left and to the right of the word and gap for which a prediction is being made; smaller values such a $C=1$ gave, however, a very similar performance.

\section{Experimental setting} \label{ss:experiments}

\subsection{Datasets provided by the organizers}
\label{ss:datasets}

Six datasets were provided for the shared task on MT QE at WMT 2018~\citep{WMT2018}, covering four language pairs ---English--German (EN--DE), German--English (DE--EN), English--Latvian (EN--LV), and English--Czech (EN--CS)--- and two MT systems ---statistical MT (SMT) and neural MT (NMT). Each dataset is split into training, development and test sets. From the data provided by the organizers of the shared task, the approach in this paper used:
\begin{enumerate} \setlength{\itemsep}{-2pt}
 \item \label{it:s} set of segments $S$ in source language,
 \item \label{it:t} set of translations $T$ of the SL segment produced by an MT system,
 \item \label{it:wg} word-level MT QE gold predictions for each word and gap in each translation $T$, and
 \item \label{it:baseline} baseline features\footnote{\url{https://www.quest.dcs.shef.ac.uk/quest_files/features_blackbox_baseline_17}} for word-level MT QE.
\end{enumerate}

Regarding the baseline features, the organizers provided 28 features per word in the dataset, from which we only used the 14 numeric features plus the part-of-speech category (one-hot encoded). This was done for the sake of simplicity of our architecture. It is worth mentioning that no valid baseline features were provided for the EN--LV datasets. In addition, the large number of part-of-speech categories in the EN--CS dataset led us to discard this feature in this case. As a result, 121 baseline features were obtained for EN--DE (SMT), 122 for EN--DE (NMT), 123 for DE-EN (SMT), 14 for EN--CS (SMT), and 0 for EN--LV (SMT) and EN--LV (NMT).

\subsection{External bilingual resources}
\label{ss:OPUS}

As described above, our approach uses ready-made, publicly available
phrase tables as bilingual resources. 
In particular, we have used the
cleaned phrase tables available on June 6, 2018 in OPUS for the
language pairs involved. 
These phrase tables were built on a corpus of about 82 million pairs of sentences for DE--EN, 7 million for EN--LV, and 61 million for EN--CS.
Phrase tables were available only for one translation direction and some of them had to be inverted (for example, in the case of EN--DE or EN--CS).



\section{Results} \label{se:results}

\begin{table*}[t!]
\begin{center}
\begin{tabular}{|l|rrr|r|rrr|rrr|}
\hline & \multicolumn{4}{c|}{\bf sentence-level} & \multicolumn{3}{c|}{\bf word-level (words)} & \multicolumn{3}{c|}{\bf word-level (gaps)} \\ \cline{2-11}
 \bf Dataset & \bf $r$ & \bf MAE & \bf RMSE & \bf $\rho$ & \bf $F_\mathrm{BAD}$ & \bf $F_\mathrm{OK}$ & \bf $F_\mathrm{MULTI}$ & \bf $F_\mathrm{BAD}$ & \bf $F_\mathrm{OK}$ & \bf $F_\mathrm{MULTI}$ \\ \hline
EN--DE SMT & 0.45 & 0.15 & 0.19 & 0.44 & 0.35 & 0.81 & 0.29 & 0.33 & 0.96 & 0.32 \\
EN--DE NMT & 0.35 & 0.14 & 0.20 & 0.41 & 0.22 & 0.86 & 0.19 & 0.12 & 0.98 & 0.12 \\
DE--EN SMT & 0.63 & 0.12 & 0.17 & 0.60 & 0.43 & 0.87 & 0.37 & 0.33 & 0.97 & 0.32 \\
EN--LV SMT & 0.36 & 0.20 & 0.26 & 0.34 & 0.27 & 0.82 & 0.22 & 0.15 & 0.94 & 0.14 \\
EN--LV NMT & 0.56 & 0.17 & 0.22 & 0.55 & 0.44 & 0.80 & 0.36 & 0.17 & 0.95 & 0.16 \\
EN--CS SMT & 0.43 & 0.18 & 0.23 & 0.46 & 0.42 & 0.75 & 0.31 & 0.15 & 0.95 & 0.15 \\
\hline
\end{tabular}
\end{center}
\caption{\label{tb:result} Results for sentence-level MT QE (columns 2--5) in terms of the Pearson's correlation $r$, MAE, RMSE, and Sperman's correlation $\rho$ (for ranking). Results for the task of word labelling (columns 6--8) and gap labelling (columns 9--11) in terms of the $F_1$ score for class \textit{BAD} ($F_\mathrm{BAD}$), the $F_1$ score for class \textit{OK} ($F_\mathrm{OK}$) and the product of both ($F_\mathrm{MULTI}$).}
\end{table*}

This section describes the results obtained by the UAlacant system in the MT QE shared task at WMT 2018~\citep{WMT2018}, which are reported in Table~\ref{tb:result}. Our team participated in two sub-tasks: sentence-level MT QE (task 1) and word-level MT QE (task 2). For sentence-level MT QE we computed the number of word-level operations predicted by our word-level MT QE approach and normalized it by the length of each segment $T$, in order to obtain a metric similar to TER. The words tagged as \textit{BAD} followed by gaps tagged as \textit{BAD} were counted as replacements, the rest of words tagged as \textit{BAD} were counted as deletions, and the rest of gaps tagged as \textit{BAD} were counted as one-word insertions.\footnote{Note that this approach is rather limited, as it ignores block shifts and the number of words to be inserted in a gap, which are basic operations to compute the actual TER value.} This metric was used to participate both in the scoring and ranking sub-tasks. 

Columns 2 to 5 of Table~\ref{tb:result} show the results obtained for task 1 in terms of the Pearson's correlation $r$ between predictions and actual HTER, mean average error (MAE), and root mean squared error (RMSE), as well as Sperman's correlation $\rho$ for ranking.

Columns 6 to 11 show the results for task 2 in terms of $F_1$ score both for categories \textit{OK} and \textit{BAD},\footnote{For word deletion identification, a word marked as \emph{BAD} means that the word needs to be deleted, while in the case of insertion position identification, if a gap is marked as \textit{BAD} it means that one or more words need to be inserted there.} together with the product of both $F_1$ scores, which is the main metric of comparison of the task. The first three columns contain the results for the sub-task of labelling words while the last three columns 9 to 11 contain the results for the sub-task of labelling gaps.

As can be seen, the best results were obtained for the language pair DE--EN (SMT). Surprisingly, the results obtained for EN--LV (NMT) were also specially high for word-level and sentence-level MT QE. These results for the latter language pair are unexpected for two reasons: 
first, because no baseline features were available for word-level MT QE task for this language pair, and second, because the size of the parallel corpora from which phrase tables for this language pair were extracted were an order of magnitude smaller. One may think that the coverage of machine translation by the phrase tables could have an impact on these results. To confirm this, we checked the fraction of words in each test set that were not covered by any sub-segment pair $(\sigma,\tau)$. This fraction ranges from 15\% to 4\% depending on the test set, and has the lowest value for EN--LV (NMT); however, it is not clear that a higher coverage always leads to a better performance as one of the datasets with a better coverage was EN--LV (SMT) (5\%) which, in fact, obtained the worst results in our experiments.

It is worth noting that, when looking at the results obtained by other participants, the differences in performance between the different datasets seems to be rather constant, showing, for example, a drop in performance for EN--DE (NMT) and EN--LV (SMT); this lead us to think that the test set might be more difficult in these cases. One thing that we could confirm is that, for these two datasets, the ratio of  \textit{OK}/\textit{BAD} samples for word-level MT QE is lower, which may make the classification task more difficult.

In comparison with the rest of systems participating in this task, UAlacant was the best-performing one in the sub-task of labelling gaps for 3 out of the 6 datasets provided (DE--EN SMT, EN--LV SMT, and EN--LV NMT). Results obtained for the sub-task of labelling words were poorer and usually in the lower part of the classification. However, the sentence-level MT QE submissions, which build on the labels predicted for words and gaps by the word-level MT QE system, performed substantially better and outperformed the baseline for all the datasets but EN--DE (NMT) and, for EN--LV (NMT), it even ranked third.

As said above, one of the main advantages of this approach is that it can be trained with limited computational resources. In our case, we trained our systems on a AMD Opteron(tm) Processor 6128 CPU with 16 cores and, for the largest set of features (dataset DE--EN SMT), training took 2,5 hours, about 4 minutes per epoch.\footnote{Total training time corresponds to 35 epochs.}

\section{Concluding remarks} \label{se:conclusion}

We have presented a simple MT word-level QE method that matches the content of publicly available statistical MT phrase pairs to the source segment $S$ and its machine translation $T$ to produce a number of features at each word and gap. To predict if the current word has to be deleted or if words have to be inserted in the current gap, the features for the current word and gap and $C$ words and gaps  to the left and to the right are processed by a two-hidden-layer feed-forward NN. When compared with other participants in the WMT 2018 shared task, our system ranks first in labelling gaps for 3 of the 6 language pairs, but does not perform too well in labelling words. We also used word-level estimations to approximate TER. We participated with this approximation in the sentence-level MT QE sub-task obtaining a reasonable performance ranking, for almost all datasets, above the baseline.

One of the main advantages of the work presented here is that it does not require huge computational resources, and it can be trained even on a CPU in a reasonable time.

\section*{Acknowledgments}
Work funded by the Spanish Government through the EFFORTUNE project (project number TIN\-2015-69632-R).

\bibliography{emnlp2018}
\bibliographystyle{acl_natbib_nourl}

\end{document}